\documentclass[10pt,twocolumn,letterpaper]{article}

\usepackage{cvpr}              %

\usepackage[dvipsnames]{xcolor}
\usepackage{graphicx}

\usepackage{algorithm}  
\usepackage{algorithmicx}
\usepackage{algpseudocode}
\usepackage{multirow}
\usepackage{color, colortbl}
\newlength\savewidth\newcommand\shline{\noalign{\global\savewidth\arrayrulewidth
  \global\arrayrulewidth 1pt}\hline\noalign{\global\arrayrulewidth\savewidth}}

\definecolor{Gray}{gray}{0.95}

\definecolor{cvprblue}{rgb}{0.21,0.49,0.74}
\usepackage[pagebackref,breaklinks,colorlinks,citecolor=cvprblue]{hyperref}

\title{Stitched ViTs are Flexible Vision Backbones}

\author{Zizheng Pan  \quad Jing Liu \quad Haoyu He \quad Jianfei Cai \quad Bohan Zhuang\textsuperscript{$\dagger$} \\[0.2cm]
ZIP Lab, Monash University, Australia
}

\begin{document}

\twocolumn[{
\maketitle
\begin{figure}[H]
\hsize=\textwidth
\centering
    \includegraphics[width=\textwidth]{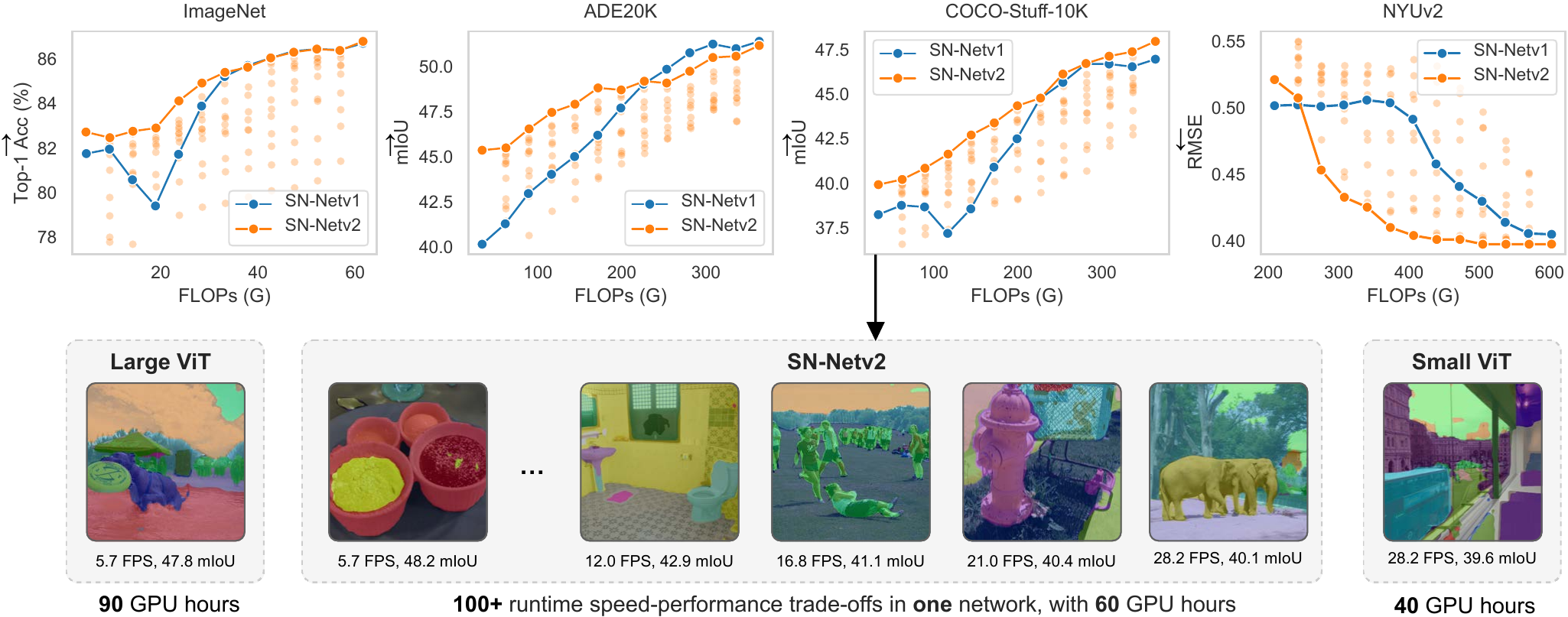}
    \caption{We adapt the framework of stitchable neural networks (SN-Net) into \textbf{downstream dense prediction tasks}. Compared to SN-Netv1 \cite{pan2023snnets}, the new framework consistently improves the performance at low FLOPs while maintaining competitive performance at high FLOPs across different datasets, thus obtaining a better Pareto frontier (highlighted in lines). Compared to individual large ViT trainings, SN-Netv2 obtains numerous trade-offs with less A100 GPU training hours. At the inference time, SN-Netv2 inherits the advantage of SN-Net which allows \textbf{runtime speed-performance trade-off switching} to satisfy different resource constraints. Besides ImageNet-1K classification, we evaluate the performance of semantic segmentation in mIoU on ADE20K and COCO-Stuff-10K, and depth estimation in RMSE on NYUv2.
    }
    \label{fig:banner}	
    \end{figure}
}]

\footnotetext{\textsuperscript{$\dagger$}Corresponding author. E-mail: $\tt  bohan.zhuang@gmail.com$}

\begin{abstract}
  Large pretrained plain vision Transformers (ViTs) have been the workhorse for many downstream tasks. However, existing works utilizing off-the-shelf ViTs are inefficient in terms of training and deployment, because adopting ViTs with individual sizes requires separate trainings and is restricted by fixed performance-efficiency trade-offs. In this paper, we are inspired by stitchable neural networks (SN-Net), which is a new framework that cheaply produces a single model that covers rich subnetworks by stitching pretrained model families, supporting diverse performance-efficiency trade-offs at runtime. Building upon this foundation, we introduce SN-Netv2, a systematically improved model stitching framework to facilitate downstream task adaptation. Specifically, we first propose a two-way stitching scheme to enlarge the stitching space. We then design a resource-constrained sampling strategy that takes into account the underlying FLOPs distributions in the space for better sampling. Finally, we observe that learning stitching layers as a low-rank update plays an essential role on downstream tasks to stabilize training and ensure a good Pareto frontier. With extensive experiments on ImageNet-1K, ADE20K, COCO-Stuff-10K and NYUv2, SN-Netv2 demonstrates superior performance over SN-Netv1 on downstream dense predictions and shows strong ability as a flexible vision backbone, achieving great advantages in both training efficiency and deployment flexibility. Code is available at \url{https://github.com/ziplab/SN-Netv2}.
\end{abstract}

\section{Introduction} \label{sec:intro}
General-purpose Transformer architectures~\cite{gpt3,fang2022eva,vit22b,zhai2022scaling,sam,dino} have grown into unprecedented scale in recent research. In computer vision, large pretrained plain ViTs such as MAE~\cite{mae}, DINO~\cite{dino,dinov2} and DeiT~\cite{deit,deit3} are widely adopted as backbones for tackling downstream tasks. However, despite the great performance, when adapting large pretrained ViTs to the downstream tasks, they face the challenge of huge computational cost. For example, for semantic segmentation on COCO-Stuff-10K, DeiT3-Large~\cite{deit3} suffers from longer fine-tuning time (90 GPU hours) and slow inference speed (5.7 FPS), compared to DeiT3-S (40 GPU hours and 28.2 FPS), see Fig.~\ref{fig:banner}.

On the other hand, most existing efforts adopting pretrained ViTs as backbones for downstream tasks can be roughly categorised into three types of approaches: full finetuning~\cite{vitdet,setr,segmenter,MIMDet}, parameter-efficient finetuning~\cite{vpt,adaptformer} and adapters~\cite{chen2022vitadapter}. 
Specifically, full finetuning trains one scale of ViTs on downstream tasks
while parameter-efficient methods additionally reduce the number of learnable parameters. Adapter methods like ViT-Adapter~\cite{chen2022vitadapter} equip plain ViTs with pyramid feature maps for dense predictions. 
However, these methods lack deployment flexibility since they cannot cope with diverse deployments with different computing budget requirements.%

\begin{figure}[]
	\centering
	\includegraphics[width=\linewidth]{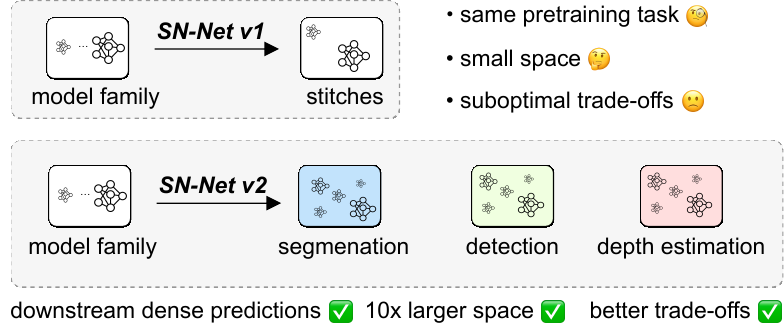}
	\caption{Comparison between SN-Netv1 and SN-Netv2. Different from SN-Netv1 which only considers the classification task on the same data domain of the pretrained model family, SN-Netv2 solves the limitations of SN-Netv1 in downstream tasks and successfully extends the practical usage of the framework.}
	\label{fig:banner_1}
    \vspace{-5pt}
\end{figure}

\begin{figure*}[]
    \vspace{-10pt}
    \centering
    \includegraphics[width=\linewidth]{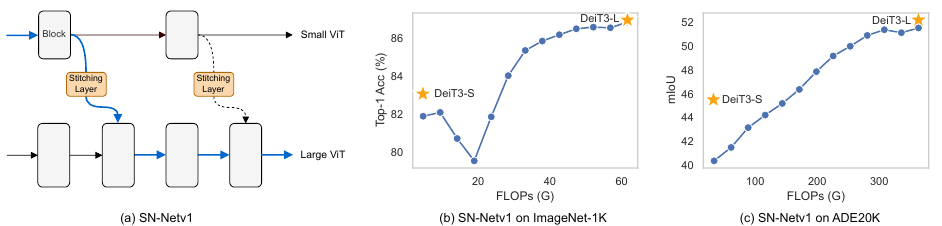}
    \caption{Explanation of SN-Netv1~\cite{pan2023snnets}. \textbf{Figure (a)}: The framework of SN-Netv1 under the default Fast-to-Slow strategy, where the blue line indicates the forward path of a stitched network by stitching a small ViT to a large ViT. \textbf{Figure (b)}: The naive Fast-to-Slow stitching strategy results in sub-optimal trade-offs in the low-FLOPs range on the pretrained dataset. \textbf{Figure (c)}: Directly applying SN-Netv1 into ADE20K obtains a bad frontier, especially when comparing with the individually trained small ViT (40.4 vs. 45.5 mIoU).}
    \label{fig:snnet_baseline}
    \vspace{-10pt}
\end{figure*}

Recently, a flexible deployment framework named
stitchable neural network (SN-Netv1~\cite{pan2023snnets}) has been proposed, which leverages a pretrained model family to obtain numerous sub-networks with different efficiency-performance trade-offs. A simple illustration is shown in Fig.~\ref{fig:snnet_baseline} (a), where SN-Netv1 inserts a few stitching layers (1 $\times$ 1 convolutional layers) between two scales of pretrained ViTs (\ie, ``anchors'' in SN-Netv1). Next, SN-Netv1 adopts a Fast-to-Slow strategy which forwards the activations from the small ViT into the large ViT via different stitching layers, thus obtaining many sub-networks (\ie, ``stitches''). With a few epochs of finetuning on the same pretraining dataset as anchors, SN-Netv1 can instantly switch network topology at runtime under various resource constraints.

Despite offering a flexible deployment framework, SN-Netv1 suffers from several major limitations. First, SN-Netv1 performs full finetuning on the same original dataset as the anchors. This might not be practical since many general-purpose ViTs are trained on exceptionally large datasets (\eg, JFT~\cite{zhai2022scaling}, LAION~\cite{laion_5b}). Second, SN-Netv1 only investigates image classification, whose data domain is different from downstream task data domains. Third, due to the small stitching space, SN-Netv1 can produce sub-optimal performance-efficiency trade-offs, and the domain gap can lead to more severe situations on downstream dense predictions, as shown in Fig.~\ref{fig:snnet_baseline} (b) and (c). 

In this paper, we introduce a new framework called \textbf{SN-Netv2}, serving as a strong and flexible vision backbone on downstream vision tasks, which systematically improves SN-Netv1 for
\textit{direct adaptation on downstream dense predictions to obtain a smooth and high-performance curve across all FLOPs (see Fig.~\ref{fig:banner} and Fig.~\ref{fig:banner_1}}). 
Specifically, 
we propose three tightly coupled improvements:
1) Different from SN-Netv1 that only permits the stitching direction of Fast-to-Slow, we propose a novel Two-way Stitching (TWS) strategy (Fig.~\ref{fig:dual_stitching}), which enables stitching to go both Fast-to-Slow and Slow-to-Fast, or travel between anchors as a round trip (\eg, Fast-Slow-Fast). As a result, we effectively enlarge the stitching space by $10\times$ and avoid sub-optimal stitches that reside in certain resource constraints.
However, under the default random sampling strategy, simply enlarging the space incurs unbalanced training and hinders the overall performance due to the varying number of stitches residing on different FLOPs intervals.
2) To this end, we further introduce a Resource-constrained Sampling (ROS) strategy, which draws a stitch at each iteration according to the categorical distribution of stitches in the space~\cite{attn_nas,liu2022focusformer} so as to ensure a balanced training for stitches under different resource constraints.
3) Last, %
to address the training instability caused by the domain gap between anchors and the target task, along with the increased interference among stitches in the enlarged stitching space, %
unlike the full-finetune strategy of SN-Netv1, we propose to stabilize training with a low-rank adaption of stitching layers (LoRA SL), which ensures a smooth performance curve. 
Note that we utilize LoRA as a novel solution for the critical training challenge of adapting SN-Net to new downstream tasks, which is different from exploring LoRA~\cite{lora} for parameter-efficient finetuning in existing methods~\cite{adaptformer,compacter}.

With the improved techniques, we comprehensively experiment with SN-Netv2 on ImageNet-1K, ADE20K, COCO-Stuff-10K, NYUv2 and COCO-2017. We show SN-Netv2 demonstrates stronger performance and better training efficiency than the typical single-scale ViT backbone adoption on downstream dense prediction tasks. Particularly, on both the pretraining and downstream tasks, SN-Netv2 consistently outperforms SN-Netv1 at low FLOPs, while keeping competitive performance at high FLOPs. 
Moreover, on downstream dense predictions, SN-Netv2 achieves competitive performance with anchors under \textit{equal} training schedules, while eliminating the need to train different scales of ViT backbones separately.

Our main contributions can be summarized as follows:
\begin{itemize}[]
    \item Compared to SN-Netv1 which only considers the classification task on the same data domain as pretrained models, we explore the applicability of SN-Net in downstream dense predictions. 
    \item To overcome the suboptimal trade-offs inherent in SN-Netv1 in downstream tasks, we propose SN-Netv2, which seamlessly integrates three tightly coupled components, TWS, ROS and LORA SL, and achieves a superior Pareto frontier in terms of performance and efficiency.
    \item We conduct comprehensive experiments across various datasets and tasks to evaluate SN-Netv2, showcasing its great potential as a strong and flexible vision backbone 
for downstream vision tasks.
\end{itemize}

\section{Related Work} \label{sec:related_works}

\textbf{General-purpose Transformers.}
Benefit from the large-scale datasets~\cite{laion_5b,zhai2022scaling} and powerful scaling laws~\cite{kaplan2020scaling}, recent pretrained plain ViTs~\cite{mae,beit,vit,vit22b} have achieved strong performance on many visual benchmarks. Based on different training objectives, ViT pretraining can be done by either supervised learning (SL) or self-supervised learning (SSL). In SL-based pretraining, the common practice~\cite{swin,vit} is to train ViTs with the cross-entropy loss on ImageNet-21K~\cite{imagenet} or privately gathered images~\cite{sun2017revisiting}. Compared to it, SSL-based pretraining is a more promising direction as it can leverage the vast amounts of unlabeled data. Prevalent approaches in this area include contrastive learning~\cite{dino,mocov3,yu2022coca,clip} and masked image modeling (MIM)~\cite{mae,beit,ibot,peco,convmae,simmim}. In recent studies, ViTs have been scaled up to billion parameters~\cite{zhai2022scaling,vit22b,fang2022eva} to match the prevalent large language models (LLMs)~\cite{opt,gpt2,gpt3}. 
However, when adopting into downstream tasks, most existing efforts~\cite{vitdet,MIMDet,segvit,setr} fully adopt the pretrained model in order to exploit the pretrained weights, suffering from the huge computational cost and memory consumption. 
In contrast, this paper efficiently adapts large pretrained ViTs into CV tasks as a single flexible backbone to satisfy various resource constraints at runtime.

\noindent\textbf{Model stitching.}
Model stitching has been studied in previous works~\cite{stitch_0, stitch_1, stitch_2} to measure the similarity of representations from neural networks. In particular, it implies a sequence of well-performed networks that can be obtained by stitching the early portion of a trained network with the last portion of another trained network by a $1\times1$ convolution layer (a.k.a, stitching layer). Inspired by this observation, Yang~\etal~\cite{dmr} proposed to dissect and reassemble a new architecture based on the model zoo. Most recently, Pan~\etal proposed SN-Net~\cite{pan2023snnets} to cheaply produce numerous networks with different complexity-performance trade-offs by stitching a family of pretrained models. However, it only experiments with the same pretraining domain, without exploring dense prediction performance. In this paper, we systematically improves SN-Net and address its limitations on the stitching space, sampling strategy and downstream training adaptation.

\section{Rethinking Stitchable Neural Networks} \label{sec:rethinking}

\begin{figure*}[]
	\centering
	\includegraphics[width=1.0\linewidth]{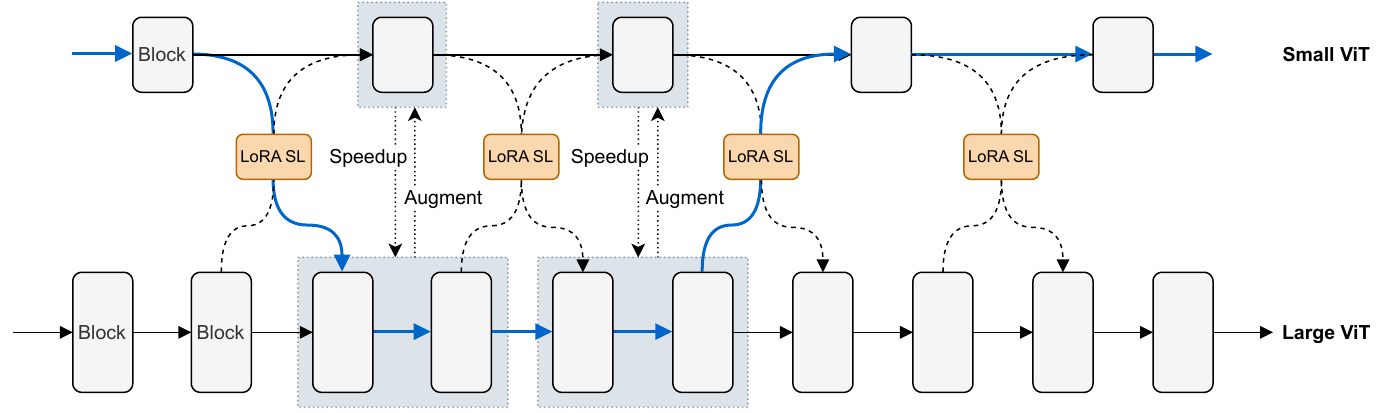}
	\caption{The framework of the proposed \textbf{Two-way Stitching}, where stitching can go Fast-to-Slow,  Slow-to-Fast, or travel between anchors as a round trip. For example, the blue line represents the forward path of a new stitched network, which starts with blocks from the small ViT, connects a few blocks from the large ViT, and finally propagates its activations back to the small ViT. ``LoRA SL'' refers to the proposed low-rank adaptation of stitching layers.}
	\label{fig:dual_stitching}
\end{figure*}

SN-Netv1~\cite{pan2023snnets} is a scalable framework built upon pretrained model families. A simple illustration of SN-Netv1 is shown in Figure~\ref{fig:snnet_baseline} (a).
Specifically, given an input $\mathbf{X}$ and two pretrained models $f_{\theta}$ and $f_{\phi}$, $\mathcal{S} : \mathcal{A}_{\theta, l} \to \mathcal{A}_{\phi,m}$ is denoted as a stitching layer which implements a transformation between the activation space of the $l$-th layer of $f_\theta$ to the activation space of the $m$-th layer of $f_\phi$. Let  $Z_{\theta}$ and $G_{\phi}$ represent the function of the first and last portion of blocks of $f_{\theta}$ and $f_{\phi}$ respectively.
Next, SN-Netv1 obtains a new network architecture $ F_{\mathcal{S}}$ by 
\begin{equation} \label{eq:stitch}
    F_S(\mathbf{X}) = G_{\phi,m} \circ \mathcal{S} \circ Z_{\theta,l}(\mathbf{X}),
\end{equation}
where $\circ $ indicates the composition. With different stitching configurations of $l$ and $m$, SN-Net cheaply produces numerous stitched networks (\ie, stitches) that achieve good accuracy-efficiency trade-offs between two pretrained models (\ie, anchors) after joint training. However, despite the simple idea and its effectiveness, SN-Net still suffers from noticeable limitations.

\noindent\textbf{Stitching space.}
SN-Netv1 strictly follows a typical stitching direction in the literature~\cite{stitch_1,stitch_2}, \ie, propagating activations by starting with one anchor and ending with another. Based on whether the stitched network starts with a small anchor (Fast) or a large anchor (Slow), Pan~\etal~\cite{pan2023snnets} explored two stitching directions: Fast-to-Slow and Slow-to-Fast, and demonstrated that Fast-to-Slow generally leads to a better and more smooth performance curve. 
However, simply adhering to the Fast-to-Slow direction may assemble sub-optimal network architectures. As shown in Figure~\ref{fig:snnet_baseline} (b), stitching DeiT3-S and DeiT3-L under the default setting of SN-Netv1 produces a few stitches that achieve worse performance than the small anchor, even with higher FLOPs. Therefore, it is evident that the existing stitching space in SN-Netv1 requires redesign.

\noindent\textbf{Sampling strategy.}
Training SN-Netv1 is simple but effective: at each training iteration, SN-Netv1 randomly samples a stitch from a pre-defined configuration set, and then train the sampled stitch as a normal network with gradient descent. However, random sampling approach only works well when the stitches are evenly distributed across different FLOPs constraints (\eg, 5G, 10G). While this condition is met in the initial stitching space of SN-Netv1, \textit{enlarging the space can result in a problematic imbalance, with some FLOPs constraints having far fewer stitches than others}. Consequently, this leads to an unbalanced training for networks in certain FLOPs ranges, and thus negatively impacts the performance.

\noindent\textbf{Stitching layers.}
SN-Netv1 is initially trained on the same data source of the pretrained model families (\eg, ImageNet-1K pretrained DeiTs~\cite{deit}). Under this setting, training a stitching layer with full finetune has a consistent target as the activation space $\mathcal{A}_{\theta, l}$ and $\mathcal{A}_{\phi,m}$ may not change significantly. However, things can be different when adopting SN-Netv1 on downstream dense prediction tasks due to the domain gap between the pretrained data and the target data. In this case, both $\mathcal{A}_{\theta, l}$ and $\mathcal{A}_{\phi,m}$ need to adapt to the target domain, making it unstable and difficult to simultaneously learn many stitches. This implies the necessity of an appropriate method for learning stitching layers on downstream tasks.

\section{Method} \label{sec:method}
In this section, we systematically introduce our improvements over SN-Netv1, including the redesign of stitching space, sampling strategy and 
stitching layers for stable adaptation to downstream dense prediction tasks.

\subsection{Two-way Stitching}
To improve the stitching space, we first propose Two-way stitching (TWS), which allows the stitching to travel in different ways including Fast-to-Slow (FS), Slow-to-Fast (SF), Fast-Slow-Fast (FSF) and Slow-Fast-Slow (SFS).
The design principle of TWS is to augment the small anchor with the blocks from a large and strong anchor, while accelerating the large anchor by adopting more efficient blocks from the small anchor. In this way, TWS can leverage the strengths of different scales of anchors. 

We illustrate our framework in Figure~\ref{fig:dual_stitching}, where it also shows a concrete example of an FSF stitch in the blue line of Figure~\ref{fig:dual_stitching}. Specifically, 
its begins with the small anchor during the forwarding pass, traverses the large anchor along the middle route, and ultimately returns to the small anchor for subsequent propagations. By enabling a broader stitching configuration, TWS enlarges the stitching space by $10\times$ compared to the initial space of SN-Netv1 (\eg, 134 \vs 13 based on DeiT3-S/L), which facilitates the discovery of more optimal architectures, as shown in Section~\ref{sec:experiments}.  Moreover, benefiting from the two new stitching configurations (\ie, FSF and SFS), SN-Netv2 produces many networks at the similar FLOPs
as each stitch can be regarded as replacing intermediate consecutive blocks at certain positions in one anchor with the blocks from another anchor.

\subsection{Resource-Constrained Sampling}

\begin{figure}[]
	\centering
	\includegraphics[width=\linewidth]{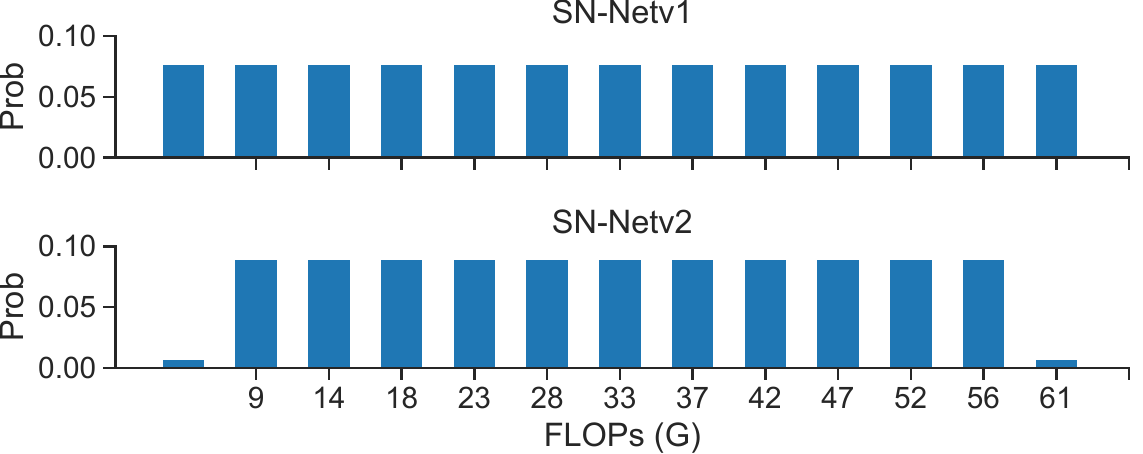}
	\caption{Comparison on the categorical distribution of stitches between SN-Netv1 and SN-Netv2 by stitching DeiT3-S and DeiT3-L on ImageNet-1K.}
	\label{fig:fos_sampling}
        \vspace{-10pt}
\end{figure}

After enlarging the stitching space, we empirically find that the performance of anchors drops significantly compared to its original performance on ImageNet. 
To understand this, we begin with analysing the categorical distribution of stitches in the space, \ie, $\pi(\tau)$, where $\tau$ denotes the FLOPs constraint. In practice, we round the real FLOPs following a step $t$ to discretize the whole FLOPs range. By default, we adopt $t=1$ in ImageNet experiments and $t=10$ for dense prediction tasks.
Benefiting from the significantly smaller architectural space compared to neural architecture search (\ie, NAS, $10^2$ \textit{vs}. $10^{20}$), 
we can calculate $\pi(\tau)$ exactly prior to training by $\pi(\tau = \tau_{0}) = \#(\tau = \tau_{0})/E$, where $E$ is the total number of stitches in the space and $\#(\tau = \tau_{0})$ is the number of stitches that yield FLOPs $\tau_{0}$. 

With the exact $\pi(\tau)$, we visualize the categorical distribution of stitches for SN-Netv2 and SN-Netv1 in Figure~\ref{fig:fos_sampling}. 
As it shows, stitches in SN-Netv1 are evenly distributed across different FLOPs, which ensures balanced training for stitches at different resource constraints. 
However, under Two-way Stitching, the sampling probability of anchors ($2/134$) is much lower than stitches in other FLOPs constraints. As a result, anchors are significantly under-trained which therefore affects the overall performance of SN-Netv2.
Inspired by recent NAS works~\cite{attn_nas,liu2022focusformer}, we design a Resource-constrained Sampling strategy (ROS) for SN-Netv2. Specifically, at each training iteration, we first sample a FLOPs constraint $\tau_{0}$. Next, we randomly sample a stitch that satisfies the constraint $\alpha \sim \pi(\tau = \tau_{0})$. With this design, ROS effectively guarantees balanced training for stitches at different FLOPs constraints, especially for anchors where we increase their sampling probability by 10$\times$, \eg, from $2/134$ to $2/13$ based on DeiT3-S/L and 13 FLOPs intervals.

\subsection{Low-Rank Adaptation of Stitching Layers}
Under Two-way stitching, the stitching layers involve with two types of transformation matrix to control how the stitches route between anchors:
$\mathbf{M}_{1} \in  \mathbb{R}^{D_1 \times D_2}$ and $\mathbf{M}_{2} \in  \mathbb{R}^{D_2 \times D_1}$, 
where $D_1$ and $D_2$ refer to model widths of the small anchor and large anchor, respectively. In practice, they are 1$\times$1 convolutions and initialized by the least-square (LS) solution as in SN-Netv1. Formally, let $\mathbf{X}_1 \in  \mathbb{R}^{N \times D_1}$ and $\mathbf{X}_2 \in \mathbb{R}^{N \times D_2}$ be the feature maps from two anchors at one stitching position, where $N$ denotes the length of the input sequence. Next, the targeted transformation matrix for $\mathcal{S}_{1}$ and  $\mathcal{S}_{2}$ can be obtained respectively by
\begin{equation} \label{eq:mp_inverse}
    \mathbf{M}_1 = \mathbf{X}_1^{\dagger}\mathbf{X}_2, \mathbf{M}_2 = \mathbf{X}_2^{\dagger}\mathbf{X}_1,
\end{equation}
where ${\dagger}$ denotes the Moore-Penrose pseudoinverse of a matrix. In SN-Netv1, the transformation matrix $\mathbf{M}$ is fully finetuned with gradient descent during training, which demonstrates better performance than the initial LS solution.

However, as mentioned in Section~\ref{sec:rethinking}, learning a good stitching layer on downstream tasks can be difficult as anchors need to adapt to the target domain simultaneously. Motivated by previous observations~\cite{fu2022effectiveness,chen-etal-2022-revisiting} that sparse finetuning via paremeter-efficient finetuning (PEFT) can impose regularization on the model which facilitates training stability, we propose LoRA SL, a low-rank adaptation method for stitching layers in order to stabilize training. Specifically, similar to LoRA~\cite{lora}, we freeze the LS initialized transformation matrix $\mathbf{M}$ but constrain its update with a low-rank decomposition. Taking $\mathcal{S}_{1}$ as a concrete example, the activation projection of the stitching layer can be formulated by 
\begin{equation} \label{eq:decompose}
\mathbf{X}_1\mathbf{M}_1 + \mathbf{X}_1\Delta\mathbf{M}_1 = \mathbf{X}_1\mathbf{M}_1 + \mathbf{X}_1\mathbf{B}_1\mathbf{A}_1,
\end{equation}
where $\mathbf{B}_1 \in  \mathbb{R}^{D_1 \times r}$, $\mathbf{A}_1 \in  \mathbb{R}^{r \times D_2}$, and $r \ll min(D_1, D_2)$ is the rank. In practice, $\mathbf{B}_1$ is initialized by Gaussian initialization and $\mathbf{A}_1$ is initialized with zeros. Therefore, the initial update $\Delta\mathbf{M}_1 = \mathbf{B}_1\mathbf{A}_1$ is zero and does not affect the LS solution at the beginning of training. 
As each stitiching layer is responsible for multiple stitches, the low-rank update helps to stabilize training. We show in Figure~\ref{fig:lora_sl_ade20k} that LoRA SL improves the overall performance of SN-Netv2.
In Algorithm~\ref{algos:training}, we summarize our training approach for SN-Netv2 and highlight the difference with SN-Netv1 in bold. It is also worth noting that we do not adopt knowledge distillation to train SN-Netv2 as it is very inefficient on downstream dense prediction tasks.

\begin{algorithm}[ht]
\caption{Training Stitchable Neural Networks v2}
    \begin{algorithmic}[1]
    \Require{Two pretrained ViTs to be stitched. \textbf{The pre-computed categorical distribution of stitches in the Two-way stitching space $\pi(\tau)$}.}
    \State{Initialize all stitching layers by least-squares matching, \textbf{freeze the weights $\mathbf{M}$ with LoRA}}.
    \For {$i = 1, ..., n_{iters}$}
        \State{Get next mini-batch of data $\mathbf{X}$ and label $\mathbf{Y}$.}
        \State{Clear gradients, \(optimizer.zero\_grad()\).}
        \State{\textbf{Sample a target resource constraint $\tau_{0}$}.}
        \State{\textbf{Randomly sample a stitch $\alpha$ that satisfies the constraint $\alpha \sim  \pi(\tau = \tau_{0})$}.}
        \State{Execute the current stitch, \(\hat{\mathbf{Y}} = F_{\alpha}(\mathbf{X})\).}
        \State{Compute loss, \(loss = criterion(\hat{\mathbf{Y}}, \mathbf{Y})\).}
        \State{Compute gradients, \(loss.backward()\).}
        \State{Update weights, \(optimizer.step()\).}
    \EndFor
    \end{algorithmic}
\label{algos:training}
\end{algorithm}

\begin{figure*}[!htb]
	\centering
	\includegraphics[width=\linewidth]{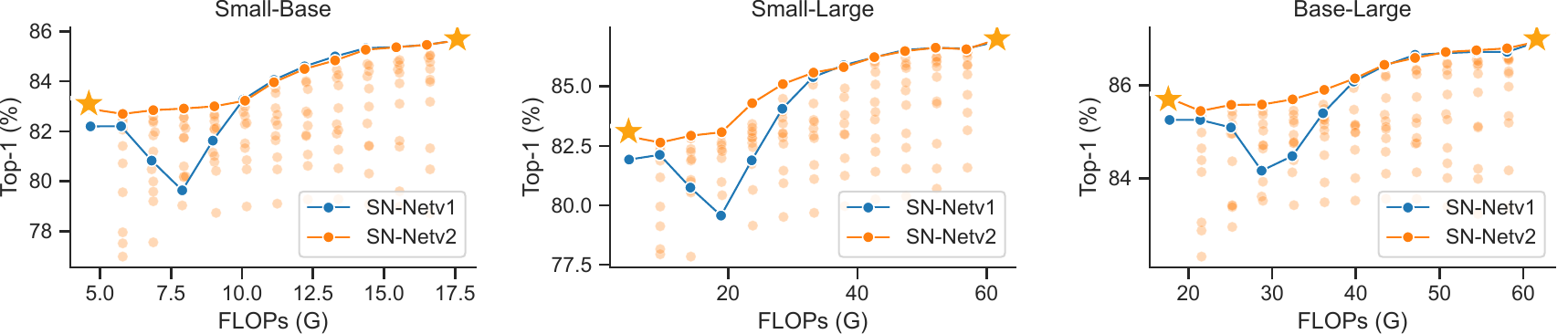}
   \vspace{-15pt}
	\caption{Performance comparison between SN-Netv1 and SN-Netv2 on ImageNet-1K based on DeiT3. The yellow stars denote the original anchor performance. We highlight the best performed stitches on the Pareto-frontier in SN-Netv2.}
  \vspace{-5pt}
	\label{fig:imagenet_res}
\end{figure*}

\begin{figure*}[htbp]
  \begin{minipage}{0.48\textwidth}
    \centering
	\includegraphics[width=\linewidth]{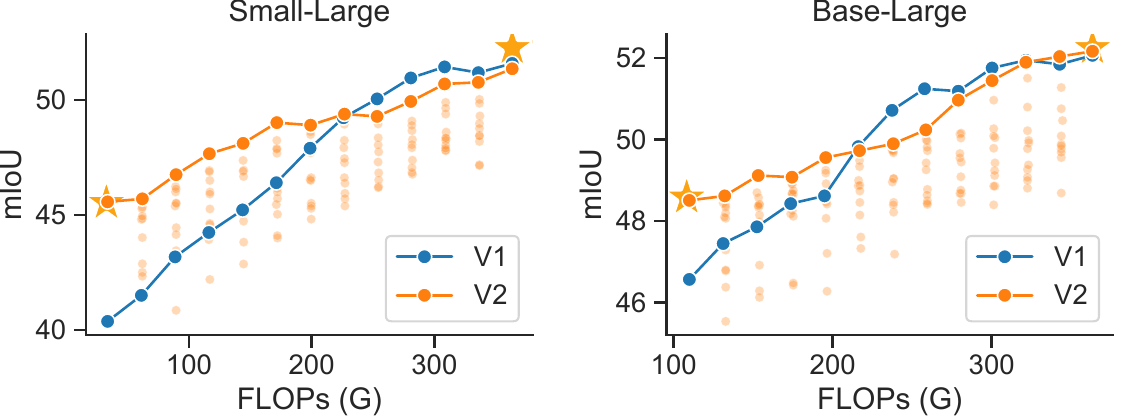}
 \vspace{-10pt}
	\caption{Semantic segmentation results of SN-Netv1 and SN-Netv2 on ADE20K by stitching DeiT3-S/L and DeiT3-B/L. We highlight the stitches at the Pareto frontier in lines.}
	\label{fig:ade20K_res}
  \end{minipage}
  \hspace{0.2cm}
  \begin{minipage}{0.48\textwidth}
    \centering
	\includegraphics[width=\linewidth]{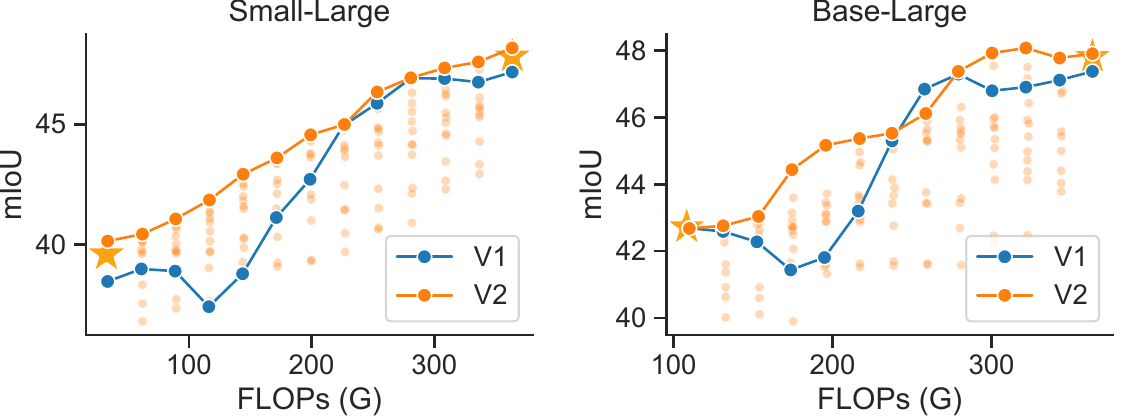}
 \vspace{-10pt}
	\caption{Semantic segmentation results of SN-Netv1 and SN-Netv2 on COCO-Stuff-10K by stitching DeiT3-S/L and DeiT3-B/L. We highlight the stitches at the Pareto frontier in lines.}
	\label{fig:coco_stuff}
  \end{minipage}
  \vspace{-10pt}
\end{figure*}

\begin{figure*}[htbp]
	\centering
	\includegraphics[width=\linewidth]{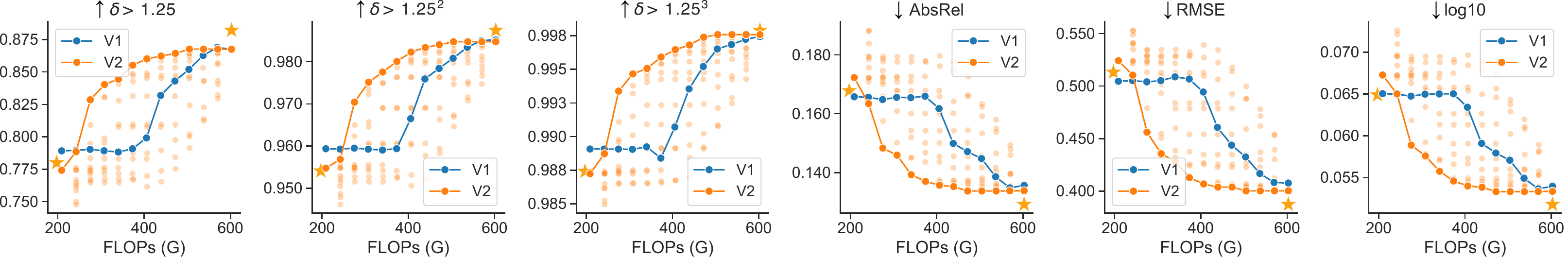}
	\caption{Results of stitching DeiT3-S and DeiT3-L on NYUv2~\cite{nyu2} under the framework of DPT~\cite{dpt}.}
 \vspace{-5pt}
	\label{fig:dpt_sl}
\end{figure*}

\section{Experiments} \label{sec:experiments}
In this section, we first show the advantage of SN-Netv2 over SN-Netv1 on ImageNet-1K~\cite{imagenet}. Next, we conduct experiments on downstream dense prediction tasks to show the strong performance of SN-Netv2 as a promising flexible vision backbone, including semantic segmentation on ADE20K~\cite{ade20k}, COCO-Stuff~\cite{coco_stuff}, and depth estimation on NYUv2~\cite{nyu2}. Due to the limited space, we provide more experiments on COCO detection in the supplement.

\subsection{ImageNet Classification}

\textbf{Implementation details.} We experiment with three combinations of stitching for DeiT3~\cite{deit3}, namely DeiT3-S/B, DeiT3-S/L and DeiT3-B/L. We train each setting on ImageNet-1K by 50 epochs with a total batch size of 256 on 8 V100 GPUs. The learning rate is 0.1$\times$ to that of default training time learning rate in DeiT3. During training, anchors adopt the same stochastic layer drop rate as the pretrained model family, \ie, 0.05, 0.15, 0.4 for DeiT3-S, DeiT3-B, DeiT3-L, respectively. All other hyperparameters are the same as in DeiT3. For all experiments, we adopt a rank of 16 for LoRA SL and 100 images for its Least-square initialization~\cite{pan2023snnets}. We evaluate the performance by Top-1 accuracy (\%).

\begin{table*}[]
\centering
\caption{Training efficiency comparison between SN-Netv2 and individual anchors based on SETR. We measure the training cost by A100 GPU hours. FLOPs and mIoU in SN-Netv2 are represented by a range, \eg, ``34 - 363'' means the model can cover FLOPs ranging from 34G to 363G.}
\vspace{-5pt}
\renewcommand\arraystretch{1.}
\label{tab:segm_training}
\scalebox{0.95}{
\begin{tabular}{l|cc|cc|cc}
\multirow{2}{*}{Model} & \multirow{2}{*}{Params (M)} & \multirow{2}{*}{FLOPs (G)} & \multicolumn{2}{c|}{ADE20K} & \multicolumn{2}{c}{COCO-Stuff-10K} \\ \cline{4-7} 
                     &     &           & mIoU        & Train Cost & mIoU        & Train Cost \\ \shline
DeiT3-S              & 23  & 32        & 45.5        &    75        & 39.6        &  40          \\
DeiT3-B              & 88  & 108       & 48.6        &    90        & 42.7        &  48          \\
DeiT3-L              & 307 & 363       & 52.3        &    174        & 47.8        & 90           \\
\rowcolor{Gray}
\textbf{SN-Netv2 + DeiT3-S/L} & 338 & \textbf{34 - 363}  & \textbf{45.6 - 51.4} &    \textbf{120}        & \textbf{40.1 - 48.2} &   \textbf{60}         \\
\rowcolor{Gray}
\textbf{SN-Netv2 + DeiT3-B/L} & 412 & \textbf{110 - 363} & \textbf{48.5 - 52.2} &    \textbf{140}        & \textbf{42.7 - 48.1} &   \textbf{80}        
\end{tabular}
}
\end{table*}

\noindent\textbf{Results.} 
In Figure~\ref{fig:imagenet_res}, we compare SN-Netv2 to SN-Netv1 on ImageNet-1K. With the same training schedule, SN-Netv2 produces hundreds of stitches that satisfy a diverse range of resource constraints. More importantly, by highlighting the stitches on the Pareto frontier, we show SN-Netv2 can find much better architectures, \ie, stitching configurations than SN-Netv1. 
This is achieved by the enlarged stitching space from Two-way stitching, the improved ROS sampling, as well as the effective low-rank update of stitching layers under LoRA SL.
Moreover, while SN-Netv1 results in a noticeable performance drop for the small anchor, SN-Netv2 escapes from the sub-optimal space at the low-FLOPs constraints and significantly improves the stitches that reside in that range. Overall, this strongly demonstrates the advantage of SN-Netv2 over SN-Netv1.

\subsection{Semantic Segmentation}
\textbf{Implementation details.} To explore the power of SN-Netv2 on downstream tasks, we first conduct comprehensive experiments on semantic segmentation, including ADE20K~\cite{ade20k} and COCO-Stuff-10K~\cite{coco_stuff}. Our method is based on SETR~\cite{setr} due to its simple framework which well reflects the performance of plain ViT backbones~\cite{segvit}. It is worth noting that while SETR is proposed along with three different decoders: Naive, PUP and MLA, we adopt the Naive approach as it achieves the best performance-efficiency trade-off. For all experiments, we train with a total batch size of 16 for ADE20K and COCO-Stuff-10K. We set the training iterations as 160K, 80K for ADE20K and COCO-Stuff-10K, respectively. Besides, we adopt a rank of 16 for LoRA SL when stitching DeiT3-S and DeiT3-L, and a rank of 4 for stitching DeiT3-B and DeiT3-L. Similar to ImageNet experiments, we use 100 images for LS initialization. All other hyperparameters are set with default choices in mmsegmentation~\cite{mmseg2020}. Following prior works~\cite{segformer,segvit,segnext}, we adopt mean Intersection over Union (mIoU) as the metric to evaluate the performance.

\begin{figure*}[]
	\centering
  \vspace{-5pt}
	\includegraphics[width=\linewidth]{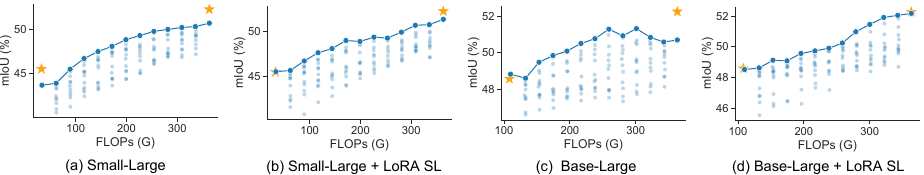}
	\caption{Performance comparison of SN-Netv2 with (Figures (b) and (d)) and without (Figures (a) and (c)) low-rank adaptation for stitching layers on ADE20K.}
	\label{fig:lora_sl_ade20k}
 \vspace{-5pt}
\end{figure*}

\noindent\textbf{ADE20K results.} We report our ADE20K results in Figure~\ref{fig:ade20K_res}. Specifically, based on DeiT3-S and DeiT3-L, SN-Netv2 demonstrates strong performance against anchor performance while simultaneously supporting a diverse range of resource constraints. By stitching DeiT3-B and DeiT3-L, SN-Netv2 achieves equal performance with anchors at their FLOPs. Moreover, by ploting the stitches on Pareto Frontier, we show SN-Netv2 smoothly interpolates the performance of two solo backbone settings. Compared to SN-Netv1, SN-Netv2 obtains a better overall Pareto frontier, demonstrating the advantage of our method.

\noindent\textbf{COCO-Stuff-10K results.} In Figure~\ref{fig:coco_stuff}, we show that stitching DeiT3-S and DeiT3-L under SN-Netv2 even achieves better performance than anchors. More impressively, by stitching DeiT3-B and DeiT3-L, we found some stitches that achieve better performance than the large anchor at a lower FLOPs. It implies that the original plain ViTs may not be the best architecture in different domains. Finally, we can observe consistently better frontiers when comparing with SN-Netv1.

\noindent\textbf{Training efficiency.} We demonstrate that SN-Netv2 achieves great training advantages compared to typical backbone adoption in downstream dense prediction tasks. As shown in Table~\ref{tab:segm_training}, on both ADE20K and COCO-Stuff-10K, stitching DeiT3-S/L or DeiT3-B/L can cover a wide range of performance-efficiency trade-offs in a single network, while requiring even less GPU hours than training the anchors separately (\eg, 140 \vs 174 + 90 on ADE20K).

\subsection{Depth Estimation} \label{sec:depth_est}

Based on DPT~\cite{dpt}, we conduct experiments on NYUv2~\cite{nyu2} dataset and train SN-Netv2 by stitching DeiT3-S/L.
We report the experiment results in Figure~\ref{fig:dpt_sl}. Overall, SN-Netv2 demonstrates highly competitive performance compared to the anchor models, while simultaneously achieving a better overall performance-efficiency curve than SN-Netv1 across different metrics. We provide implementation details in the supplementary material.

\begin{table*}[]
\centering
\vspace{-5pt}
\caption{Ablation study of each component in SN-Netv2 based on ADE20K by stitching DeiT3-S and DeiT3-L. Results are shown in mIoU. The FLOPs constraints at 34G and 363G indicate the same architecture as the DeiT3-S and DeiT3-L, respectively.}
\renewcommand\arraystretch{1.1}
\vspace{-5pt}
\label{tab:ablate_components}
\scalebox{0.825}{
\begin{tabular}{l|ccccccccccccc}
FLOPs Constraints & 34G  & 61G  & 88G  & 116G & 143G & 171G & 198G & 225G & 253G & 280G & 308G & 335G & 363G \\ \shline
SN-Netv1          & 40.4 & 41.5 & 43.2 & 44.2 & 45.2 & 46.4 & 47.9 & 49.2 & 50.0 & 51.0 & 51.4 & 51.2 & 51.6 \\
SN-Netv1 + TWS    & 42.8 & 43.3 & 44.2 & 45.4 & 46.5 & 47.8 & 48.5 & 49.3 & 49.6 & 49.6 & 49.8 & 50.2 & 50.5 \\
SN-Netv1 + TWS + ROS           & 43.7 & 43.9 & 45.5 & 46.7 & 47.5 & 48.1 & 48.8 & 49.3 & 49.8 & 50.0 & 50.2 & 50.3 & 50.7 \\
SN-Netv1 + TWS + ROS + LoRA SL & 45.6 & 45.7 & 46.8 & 47.7 & 48.1 & 49.0 & 48.9 & 49.4 & 49.3 & 49.9 & 50.7 & 50.8 & 51.4
\end{tabular}
}
\vspace{-10pt}
\end{table*}

\begin{table}[]
\centering
\caption{Comparison between full and LoRA finetunings for \textbf{the entire backbone} on COCO-Stuff10K. We measure the training speed in ``s/iter'', which is the time cost in seconds for each training iteration on 4 A100. FLOPs and mIoU in SN-Netv2 are represented as a range.}
\vspace{-5pt}
\renewcommand\arraystretch{1.1}
\label{tab:speed_compare}
\scalebox{0.9}{
\begin{tabular}{l|ccc}
Method          & FLOPs (G) & mIoU        & s/iter \\ \shline
DeiT3-L         & 363       & 47.8        & 0.95   \\
\rowcolor{Gray}
\textbf{SN-Netv2 }       & \textbf{34 - 363}  & \textbf{40.1 - 48.2} & \textbf{0.68}   \\
DeiT3-L + LoRA  & 363       & 44.8        & 0.90   \\
\rowcolor{Gray}
\textbf{SN-Netv2 + LoRA} & \textbf{34 - 363}  & \textbf{42.6 - 45.8} & \textbf{0.65 } 
\end{tabular}
}
\vspace{-13pt}
\end{table}

\subsection{Ablation Study}
\textbf{Effect of the three new components.} %
Based on DeiT3-S and DeiT3-L, we conduct experiments on ADE20K and show the ablation results in 
Table~\ref{tab:ablate_components}. 
{First}, we can see that, benefiting from TWS, SN-Netv2 finds better stitching configurations at the relatively low FLOPs (33G-198G). However, there is a performance gap at the FLOPs of DeiT3-S when compared with the individually trained one (45.5 vs. 42.8 mIoU). On ImageNet, the performance drop is more significant (-17.2\% Top-1 accuracy, shown in the supplementary material). Second, the proposed ROS reduces the performance gap at the FLOPs of the small anchor and improves the overall performance curve.
{Finally}, LoRA SL helps to further improve the performance significantly, 
achieving comparable performance with the individually trained 
DeiT3-S (45.6 vs 45.5 mIoU), while maintaining competitive performance at high FLOPs.

\noindent\textbf{Effect of LoRA SL for stabilizing training.} 
In our experiments, we empirically found that directly enlarging the space with TWS and ROS on downstream tasks results in a sub-optimal Pareto frontier, see Figure~\ref{fig:lora_sl_ade20k}.
To this end, the proposed low-rank adaptation for stitching layers is critical for ensuring the overall good performance on downstream tasks. As shown in Figure~\ref{fig:lora_sl_ade20k}, without LoRA SL, there is a noticeable performance drop at anchors when stitching DeiT3-S/L. The issue is even more pronounced when stitching DeiT3-B/L, resulting in a highly unstable performance curve. On the contrary, after applying LoRA to stitching layers, we achieve a more stable and better performance curve. We speculate that low-rank updates of stitching layers can stabilize anchor learning, thus ensuring good performance of intermediate stitches in SN-Netv2. We show the effect of different ranks in the supplementary material.

\noindent\textbf{Compare with parameter-efficient finetuning on the backbone.} SN-Netv2 is compatible with existing parameter-efficient finetuning (PEFT) techniques (\eg, LoRA~\cite{lora}) \textit{on the entire backbone} to achieve faster training speed and parameter efficiency. Here, we conduct experiments on COCO-Stuff-10K, and report the results in Table~\ref{tab:speed_compare}. First, we can see that SN-Netv2 can achieve faster training speed than the individual large ViT training while obtaining a wide range of trade-offs. This is because during training SN-Netv2 selects a stitch from different FLOPs constraints, which is more efficient than always training the same largest ViT. Second, compared to the individual large ViT training with LoRA, SN-Netv2 with LoRA achieves even better performance  (45.8 \vs  44.8), for which we conjecture SN-Netv2 may regularize the large anchor learning when combining with PEFT~\cite{fu2022effectiveness}, besides faster training speed and more trade-offs.

\section{Conclusion and Future Work} \label{sec:conclusion}
We have introduced SN-Netv2, a systematically improved training framework that effectively employs off-the-shelf foundation ViTs to obtain a flexible and strong vision backbone for downstream dense prediction tasks. 
Particularly, we proposed a two-way stitching strategy to enlarge the stitching space, %
devised a resource-constrained sampling approach to ensure balanced training for stitches that reside on different resource constraints, and introduced low-rank adaptation of stitching layers for stabilizing training on downstream tasks.
Extensive experiments across various datasets have demonstrated that SN-Netv2 achieves great training efficiency and inference flexibility than typical ViT backbone adoption, benefiting flexible model deployments in diverse real-world applications.

\noindent\textbf{Limitations.}
This work mainly aims to improve SN-Netv1 for flexible ViT adoption on downstream dense prediction tasks, which however leaves parameter efficient approaches~\cite{vpt,lora,adaptformer} under-explored. Future work may also consider a better training strategy to improve the performance of stitches on the Pareto frontier.

{
    \small
    \bibliographystyle{ieeenat_fullname}
    \bibliography{egbib}
}

\appendix

\clearpage
\begin{center}
    \Large{\textbf{Appendix}}
\end{center}

\begin{figure*}[]
	\centering
	\includegraphics[width=\linewidth]{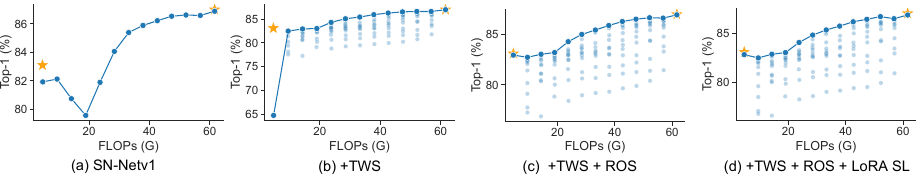}
	\caption{Effect of improvements over SN-Netv1 based on stitching DeiT3-S and DeiT3-L on ImageNet-1K.  From left to right, we gradually apply Two-way stitching (TWS), resource-constrained sampling (ROS) and low-rank adaptation of stitching layers (LoRA SL) with a rank of 16.}
	\label{fig:effect_three_improve}
\end{figure*}

\noindent We organize our supplementary material as following,
\begin{itemize}
    \item In Section~\ref{sec:eff_comp_imagenet}, we ablate our proposed three new components on ImageNet-1K, namely Two-way stitching (TWS), Resource-constrained Sampling (ROS) and LoRA Stitching Layers (LoRA SL).
    \item In Section~\ref{sec:effect_diff_ranks}, we explore the effect of different ranks in LoRA SL.
    \item In Section~\ref{sec:im_detail_depth}, we report the implementation details of depth estimation experiments on NYUv2.
    \item In Section~\ref{sec:comp_init_imgnet}, we compare different stitching types at initialization based on ImageNet-1K.
    \item In Section~\ref{sec:comp_train_imgnet}, we compare different stitching types after training based on ImageNet-1K.
    \item In Section~\ref{sec:peft_curve}, we show the detailed performance curve of applying parameter-efficient finetuning (PEFT) on the entire backbone.
    \item In Section~\ref{sec:list_previts}, we study the effect of using different pretrained ViTs in SN-Netv2.
\end{itemize}

\section{Effect of The Three New Components on ImageNet-1K} \label{sec:eff_comp_imagenet}
Based on DeiT3-S and DeiT3-L, we conduct experiments to show the effect of Two-way stitching, resource-constrained sampling and LoRA stitching layers on the pretraining dataset ImageNet-1K. As shown in Figure~\ref{fig:effect_three_improve} (b), benefiting form Two-way stitching, SN-Net successfully finds better stitching configurations at the relatively low FLOPs constraints except for the small anchor where it drops significantly. However, with the help of ROS, we improve the performance of DeiT3-S and therefore ensure the overall performance curve in Figure~\ref{fig:effect_three_improve} (c). Finally, we show LoRA SL can achieve comparable performance with the fully finetuned baseline on ImageNet-1K in Figure~\ref{fig:effect_three_improve} (d).

\section{Effect of Different Ranks in LoRA SL} \label{sec:effect_diff_ranks}
\begin{figure}[H]
	\centering
	\includegraphics[width=\linewidth]{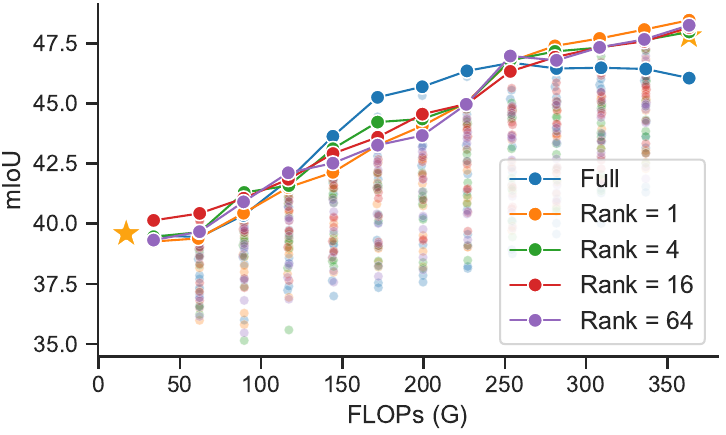}
  \vspace{-5pt}
	\caption{Effect of different ranks in LoRA SL based on stitching DeiT3-S and DeiT3-L on COCO-Stuff-10K. ``Full'' refers to fully finetune the stitching layers.}
	\label{fig:eff_diff_rank}
 \vspace{-10pt}
\end{figure}

In Figure~\ref{fig:eff_diff_rank}, we explore the effect of different ranks in LoRA SL by stitching DeiT3-S and DeiT3-L on COCO-Stuff-10K. In general, we observe that different low ranks perform similar, where they can effectively produce smoothly increasing Pareto frontiers. However, without the low-rank update, the performance of the stitches at the higher FLOPs drops ($>$250G FLOPs). Therefore, it indicates that the low-rank update may regularize stitches learning during the training time.

\begin{figure*}[]
	\centering
	\includegraphics[width=\linewidth]{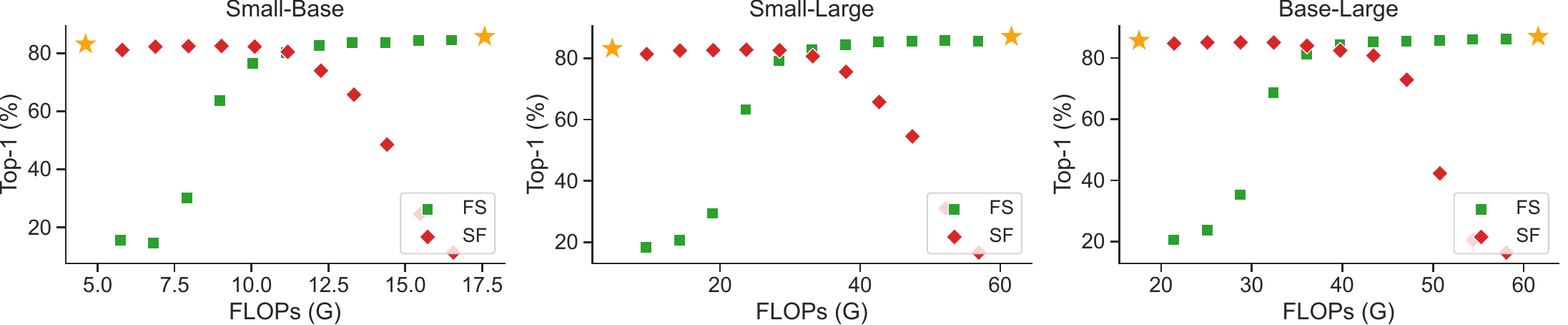}
	\caption{Performance comparison between SF and FS stitches on ImageNet-1K at initialization.}
	\label{fig:ls_init_one}
\end{figure*}

\begin{figure*}[]
	\centering
	\includegraphics[width=\linewidth]{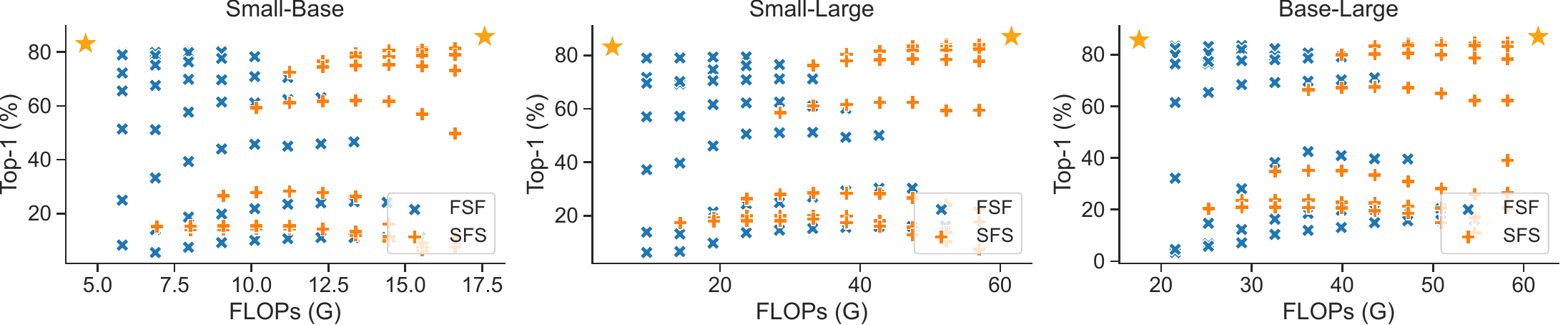}
	\caption{Performance comparison between FSF and SFS stitches on ImageNet-1K at initialization.}
	\label{fig:ls_init_two}
\end{figure*}

\begin{figure*}[]
	\centering
	\includegraphics[width=\linewidth]{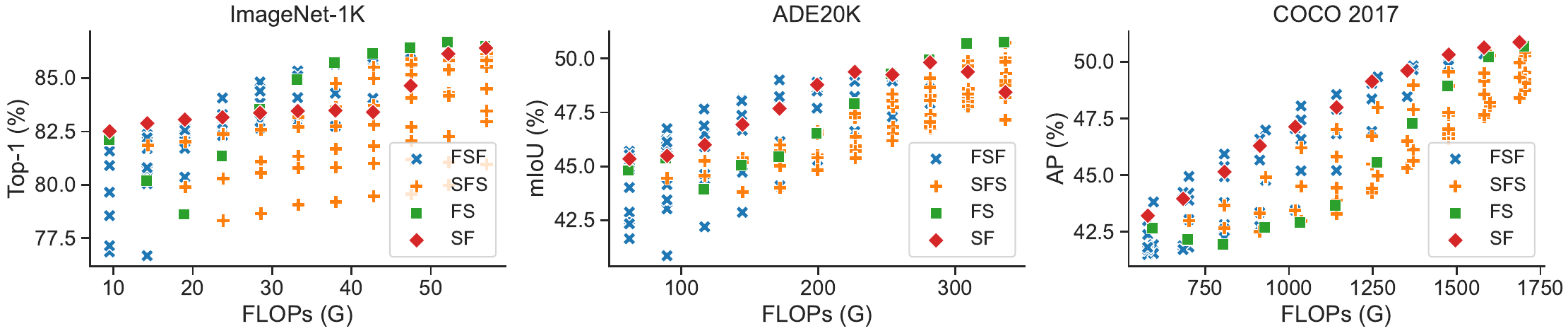}
	\caption{Ranking visualization of different types of stitches on image classification, semantic segmentation and object detection.}
	\label{fig:stitching_space}
\end{figure*}

\section{Implementation Details of Depth Estimation Experiments} \label{sec:im_detail_depth}
Based on DPT~\cite{dpt}, we conduct experiments on NYUv2~\cite{nyu2} dataset and train SN-Netv2 by stitching DeiT3-S/L and DeiT3-B/L. Specificially, we train each model on 4 A100 GPUs with a total batch size of 16. We set the training epochs to 24. We adopt AdamW optimizer with an initial learning rate of $2\times10^{-5}$. For LoRA SL, we use a rank of 4 for all experiments. We utilize common metrics for evaluting the performance on depth estimation, including $\delta > 1.25$, $\delta > 1.25^2$, $\delta > 1.25^3$ (where higher values indicate better performance), as well as AbsRel, RMSE, and Log10 (where lower values indicate better performance). All other hyperparameters are set as the default setting in DPT.

\section{Performance Comparisons of Different Stitching Types at Initialization} \label{sec:comp_init_imgnet}
By default, we randomly sample 100 images from the training set and solve the least-square problem to initialize the stitching layers. After initialization, we directly compare the performance of different types of stitches based on DeiT3 models and ImageNet-1K. 
Overall, the observations depicted in Figure~\ref{fig:ls_init_one} indicate that Fast-to-Slow (FS) outperforms at high FLOPs, while Slow-to-Fast (SF) excels at low FLOPs. This highlights the effectiveness of Two-way stitching, as no single stitching direction emerges as the superior choice across all FLOPs intervals. Similarly, in Figure~\ref{fig:ls_init_two}, we find that Fast-Slow-Fast (FSF) generally outperforms Slow-Fast-Slow (SFS) at low FLOPs, whereas SFS stitching prevails at high FLOPs. This implies that benefiting from the enlarged stitching space, SN-Netv2 can find a large number of more optimal architectures than SN-Netv1 (which uses FS only), enabling them to achieve favorable performance right from the initialization phase.

\section{Performance Comparison of Different Stitching Types after Training} \label{sec:comp_train_imgnet}
We show in Figure~\ref{fig:stitching_space} that different stitching types present varying performance under different resource constraints after training. Similar to the phenomenon in Section~\ref{sec:comp_init_imgnet}, our findings reveal that SF outperforms FS at low FLOPs, while FS becomes superior at high FLOPs, particularly for ImageNet-1K and ADE20K. We attribute this to the fact that networks adopting a few early blocks from the other anchors tend to perform better, indicating that stitching at the early stages of ViTs is more effective when stitched only once. For this, we assume early blocks involve with general low level features~\cite{HouCHBTT19,WuSH19}, which makes them more amenable to stitch. Besides, FSF stitches are generally better than that of SFS after training. These observations can serve as valuable guidelines for future stitching designs.

\begin{figure*}[]
	\centering
	\includegraphics[width=\linewidth]{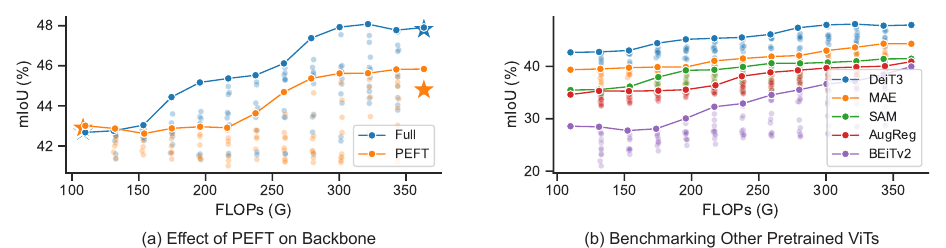}
	\caption{Figure (a): Comparing parameter-efficient tuning (PEFT) with fully finetuning (Full) on SN-Netv2. Figure (b): Benchmarking different pretrained weights of plain ViTs under SN-Netv2. Experiments are conducted on COCO-Stuff-10K and based on Base/Large variants of ViTs.}
	\label{fig:eff_pet_weights}
\end{figure*}

\section{Detailed Performance Comparison for SN-Netv2 with PEFT} \label{sec:peft_curve}
We have compared applying LoRA PEFT on the entire backbone with full finetune under SN-Netv2 in Table 3 of the main paper, where it achieves faster training speed while obtaining numerous complexity-performance trade-offs.  In this section, we report the detailed performance curve.
As shown in Figure~\ref{fig:eff_pet_weights} (a), with SN-Netv2, we achieve smooth performance-efficiency curve under PEFT with LoRA, which indicates that SN-Netv2 is compatible with PEFT techniques. It is also worth noting that compared to fully finetuning the entire backbone, PEFT has a noticeable performance drop at the high-FLOPs range. To this end, our experiments adopt full finetune on the entire backbone by default.

\section{Benchmarking Other Pretrained ViT Weights}\label{sec:list_previts}
In Figure~\ref{fig:eff_pet_weights} (b), we report the results of stitching different pretrained weights based on the base and large variants of ViTs, including MAE~\cite{mae}, SAM~\cite{sam}, AugReg~\cite{steiner2022how} and BEiTv2~\cite{beitv2}. Overall, SN-Netv2 consistently generates good Pareto frontiers for these weights. As pre-training objectives and datasets differ across ViTs, our results in Figure~\ref{fig:eff_pet_weights} (b) demonstrate varying performance when stitching different ViT weights, where DeiT3 achieves the best performance. Therefore, we choose DeiT3 as the default weights. The details of the benchmarked ViT weights are listed as following.

\begin{itemize}
    \item \textbf{DeiT3~\cite{deit3}}. ImageNet-21K pretrained DeiT3-B\footnote{\url{https://dl.fbaipublicfiles.com/deit/deit_3_base_224_21k.pth}} and DeiT3-L\footnote{\url{https://dl.fbaipublicfiles.com/deit/deit_3_large_224_21k.pth}}.
    \item \textbf{MAE~\cite{mae}}. ImageNet-1K finetuned MAE-B\footnote{\url{https://dl.fbaipublicfiles.com/mae/finetune/mae_finetuned_vit_base.pth}} and MAE-L\footnote{\url{https://dl.fbaipublicfiles.com/mae/finetune/mae_finetuned_vit_large.pth}}.
    \item \textbf{SAM~\cite{sam}}. SA-1B~\cite{sam} pretrained SAM-B\footnote{\url{https://dl.fbaipublicfiles.com/segment_anything/sam_vit_b_01ec64.pth}} and SAM-L\footnote{\url{https://dl.fbaipublicfiles.com/segment_anything/sam_vit_l_0b3195.pth}}.
    \item \textbf{AugReg~\cite{steiner2022how}}. ImageNet-21K pretrained and ImageNet-1K finetuned ViT-B\footnote{\url{https://huggingface.co/timm/vit_base_patch16_384.augreg_in21k_ft_in1k/blob/main/pytorch_model.bin}} and ViT-L\footnote{\url{https://huggingface.co/timm/vit_large_patch16_384.augreg_in21k_ft_in1k/blob/main/pytorch_model.bin}}.
    \item \textbf{BEiTv2~\cite{beitv2}}. ImageNet-1K pretrained,  then ImageNet-21K finetuned, and finally ImageNet-1K finetuned BEiTv2-B\footnote{\url{https://conversationhub.blob.core.windows.net/beit-share-public/beitv2/beitv2_base_patch16_224_pt1k_ft21kto1k.pth}} and BEiTv2-L\footnote{\url{https://conversationhub.blob.core.windows.net/beit-share-public/beitv2/beitv2_large_patch16_224_pt1k_ft21kto1k.pth}}.
\end{itemize}

\begin{figure*}[!htb]
	\centering
	\includegraphics[width=\linewidth]{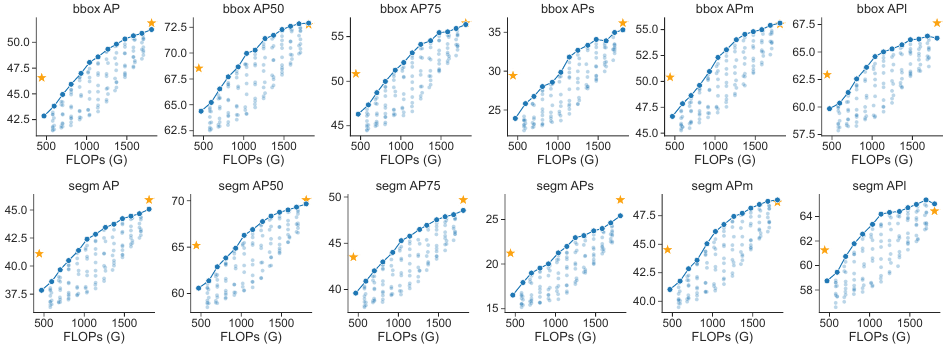}
	\caption{Object detection and instance segmentation results of SN-Netv2 on COCO-2017 by stitching DeiT3-S and DeiT3-L under Mask R-CNN~\cite{he2017mask} based ViTDet~\cite{vitdet}. We report the results on detection metrics in the top row and the instance segmentation metrics in the bottom row.}
	\label{fig:coco_res}
\end{figure*}

\section{Object Detection and Instance Segmentation}

\paragraph{Implementation details.} We experiment SN-Netv2 on COCO-2017 and adopt Mask R-CNN based ViTDet~\cite{vitdet}. We train all models including individual anchors on 8 A100 GPUs with a total batch size of 8 for 100 epochs. We set the rank as 16 for LoRA SL.
Besides, we adopt the same layer decay rate as the baselines for different anchors. All other hyperparameters adopt the default setting in detectron2~\cite{wu2019detectron2}.

\paragraph{Results.} Based on DeiT3-S and DeiT3-L, we report the performance of SN-Netv2 on object detection and instance segmentation in Figure~\ref{fig:coco_res}. Overall, we found SN-Netv2 exhibits strong flexibility on the detection task as well, as evidenced by the smooth metrics under various resource constraints.
This again demonstrates that SN-Netv2 can serve as a flexible backbone on a wide range of CV tasks. In particular, under the similar training cost ($\sim$1500 GPU hours), SN-Netv2 can achieve comparable performance at the FLOPs of individually trained DeiT3-L (51.2 \vs 51.9 bbox AP), while supporting many FLOPs-accuracy trade-offs at runtime without additional training cost.
However, we observe a performance gap between individually trained DeiT3-S with that of SN-Netv2 at the same FLOPs (42.9 \vs 46.6 bbox AP). For this, we hypothesize that with the heavy decoder in ViTDet ($27\times$ larger than SETR decoder), simultaneously ensuring the performance of hundreds of stitches as backbones can be more difficult. Nevertheless, SN-Netv2 still achieves superior advantage in terms of the training efficiency as it only requires training once while covering many FLOPs-accuracy trade-offs. We leave the improvement for future work.

\section{Memory Cost Analysis}
Intuitively, SN-Netv2 will slightly increase the memory consumption as it loads two models in GPU simultaneously. However, in practice, the additional memory overhead by loading a small ViT can be relatively minor since the main bottleneck still lies in the large ViT. For example, based on COCO-Stuff-10K and a batch size of 2, SN-Netv2 by stitching DeiT3-S and DeiT3-L results in 2,432MB peak GPU memory consumption during training, which is only 1.1$\times$ higher than individual DeiT3-L training (2,226MB). In SN-Netv2, this minor memory overhead brings a favorable trade-off for obtaining hundreds of architectures.

\end{document}